\documentclass{article}

\usepackage{aaai}

\copyrighttext{}

\usepackage[dvips]{graphicx}

\title{DES: a Challenge Problem for Nonmonotonic Reasoning Systems\thanks{%
The work of the first and third author has been funded by the Academy of Finland
(Project 43963).}
}

\author{Maarit Hietalahti \\ 
Helsinki University of Technology \\ 
Dept.\ of Computer Science and Eng. \\
Lab.\ for Theoretical Computer Science  \\
P.O.Box 5400, 02015 HUT, Finland \\
email: Maarit.Hietalahti@hut.fi 
\And 
Fabio Massacci \\
Universit\`a di Siena \\
Dipartimento di Ingegneria \\
dell'Informazione\\
via Roma 56, I-53100 Siena, Italy \\
email: massacci@dii.unisi.it
\And 
Ilkka Niemel\"a \\
Helsinki University of Technology \\ Dept.\ of Computer Science and Eng. \\
Lab.\ for Theoretical Computer Science \\
P.O.Box 5400, 02015 HUT, Finland \\
email: Ilkka.Niemela@hut.fi 
}

\newtheorem{theorem}{Theorem}

\newtheorem{definition}[theorem]{Definition}
\newtheorem{example}[theorem]{Example}

\newcommand{\eqv}{\leftrightarrow} 
\newcommand{\xor}{\oplus}

\newcommand{\lpnot}{\mathrm{not}\;} \newcommand{\lparrow}{\leftarrow}

\newcommand{\atom}[1]{p_{#1}}

\newcommand{\reduct}[2]{#1^{#2}}

\newcommand{\IFFt}{iff } 
\newcommand{\putchk}{---}

\newcommand{\inlab}[1]{\textrm{(r.#1)}}
\newcommand{\refinlab}[1]{\textrm{r.#1}}

\newcommand{\smodels}{\texttt{smodels}}
\newcommand{\lparse}{\texttt{lparse}}

\def\SMODELS{{\sf Smodels}} \def\RELSAT{{\sf rel\_sat}} 
\def\translate#1{\ensuremath{P_{#1}}}

\renewcommand{\eqv}{\Leftrightarrow}

\begin{document}

\date{} \maketitle

\begin{abstract} 
  The US Data Encryption Standard, DES for short, is put forward as an
  interesting benchmark problem for nonmonotonic reasoning systems because (i)
  it provides a set of test cases of industrial relevance which shares
  features of randomly generated problems and real-world problems, (ii) the
  representation of DES using normal logic programs with the stable model
  semantics is simple and easy to understand, and (iii) this subclass of logic
  programs can be seen as an interesting special case for many other
  formalizations of nonmonotonic reasoning.
  In this paper we present two encodings of DES as logic
  programs: a direct one out of the standard specifications and
  an optimized one extending the work of Massacci and Marraro. 
  The
  computational properties of the encodings are studied by using them for DES
  key search with the \SMODELS\ system as the implementation of the stable
  model semantics. Results indicate that the encodings and \SMODELS\ are quite
  competitive: they outperform state-of-the-art SAT-checkers working with an
  optimized encoding of DES into SAT and are comparable with a SAT-checker
  that is customized and tuned for the optimized SAT encoding.
\end{abstract}

\section{Introduction} \label{sec:introd}

Efforts on developing implementations of nonmonotonic reasoning systems have
intensified during the last years and, in particular, implementation
techniques for declarative semantics of logic programs (e.g., stable model and
well-founded semantics) have considerably advanced. With an increasing number
of systems the question of suitable test suites arises. Typical benchmarks
used for testing and comparing such systems include problems from graph
theory, planning, and constraint
satisfaction~\cite{CMMT95,DNK97,Niemela99:amai}.  However, it is still
difficult to find benchmark suites of wide industrial relevance.

In this paper we advocate that \emph{logical cryptanalysis} is a good
benchmark for nonmonotonic reasoning systems. Logical cryptanalysis has been
introduced by Massacci and Marraro~\shortcite{mass-marr-00-JAR} as a framework for
reasoning about cryptographic algorithms. They pointed out that encoding
cryptographic problems as SAT problems might be beneficial for the automated
reasoning community as it provides a set of problems of industrial relevance
which optimally shares features of randomly generated problems and real-world
problems.  Indeed, the encoding of the US Data Encryption Standard (DES) into
SAT proposed in \cite{mass-marr-00-JAR,mass-99-IJCAI} has a number of useful features:
\begin{itemize}
\item it allows to generate random instances of similar structure in
  practically inexhaustible number;
\item it provides solved instances (for which one solution is known
  beforehand) which are very hard, for which we can change the value of the
  solution, and such that we can generate as many different (hard)
  instances as we want with the same solution;
\item it has a lot of structure, and the structure is very common to many
  similar problems in hardware verification, planning and constraint
  programming (all-diff constraints, defined variables, layered definitions
  etc.).
\end{itemize}

These considerations apply to the encoding of cryptographic problems for
nonmonotonic reasoning systems with some further advantages:
\begin{itemize}
\item the representation of cryptographic algorithms using normal logic
  programs with the stable model semantics is extremely simple and easy to
  understand;
\item normal logic programs with the stable model semantics can be seen as an
  interesting special case for many other more general formalizations of
  nonmonotonic reasoning.
\end{itemize}

Indeed, we can provide a natural encoding of DES out of the
standard specifications~\cite{DES,Schn-94,Stinson98} as a logic program.
Massacci and Marraro~\cite{mass-marr-00-JAR} have developed a SAT-encoding of
DES where substantial amount of preprocessing and optimizations are employed.
As an alternative encoding of DES using logic programs we have upgraded
Massacci and Marraro's optimized SAT-encoder to deal directly with logic
programs. Using these encodings one can perform most\footnote{To be precise
  the verification of cryptographic properties proposed in
  \cite{mass-marr-00-JAR} are expressed as quantified boolean formulae. These
  are out of our scope.} of the reasoning tasks suggested
in~\cite{mass-marr-00-JAR}.

We examine the efficiency of the encodings by using an implementation of the
stable model semantics, the \SMODELS\ system~\cite{NS97:lpnmr,smodels2}, for
DES key search and by comparing the performance to that of SAT-solvers which
use the optimized encoding of DES into SAT developed Massacci and Marraro.

The rest of the paper is organized as follows. We start by briefly introducing
the stable model semantics and by discussing how to encode boolean expressions
as logic programs. We first describe the direct encoding of DES to logic
programs and then the optimized encoding. We finish with some experimental
results.

\section{Logic Programs and Stable Models} \label{sec:prelim}

The stable model semantics~\cite{GL88} generalizes the
minimal model semantics of definite programs to normal logic program rules
\begin{equation}
A \lparrow B_1,\ldots,B_m,\lpnot C_{1},\ldots,\lpnot C_n
\label{eq:lprule}
\end{equation}
where negative body literals ($\lpnot C_i$) are allowed.  For a ground
(variable-free) program $P$, the stable models are defined as follows.  The
{\em reduct} $\reduct{P}{S}$ of a program $P$ with respect to a set of atoms
$S$ is the program obtained from $P$ by deleting
\begin{enumerate}
\item each rule that has a negative literal $\lpnot C$ in its body with
  $C \in S$ and
\item all negative literals in 
the remaining rules.
\end{enumerate}
The reduct $\reduct{P}{S}$ can be seen as the set of potentially
applicable rules given the stable model $S$, i.e., as the rules where
the negative body literals are satisfied by the model.  Note that in the
reduct the negative body literals of the potentially applicable rules
are removed and, hence, the rules are definite.  The idea is that a
stable model should be grounded (or justified) in the sense that every
atom in the model is a consequence of the potentially applicable rules
and every consequence of the potentially applicable
rules is included in the model.  The atomic consequences of a set of
definite rules can be captured by the unique minimal model, the
\emph{least model}, of the set seen as definite clauses.  Hence,
a set of atoms is a stable model of a program if it coincides with the
least model of the reduct.

\begin{definition}
  Let $P$ be a ground program. Then a set of ground atoms $S$ is a
  \emph{stable model} of $P$ \IFFt $S$ is the least model of 
  $\reduct{P}{S}$.
\end{definition}

\begin{example}
\label{ex:smodel:intro}
Program $P$
\[
P:
\begin{array}[t]{l}
p \leftarrow \lpnot q, r\\
q \leftarrow \lpnot p \\
r \leftarrow \lpnot s \\
s \leftarrow \lpnot p
\end{array}
\]
has a stable model $S = \{r,p\}$ because $S$ is the least model of
$\reduct{P}{S}$. 
\[
\reduct{P}{S}:
\begin{array}[t]{l}
p \leftarrow r \\
r \leftarrow
\end{array}
\]
In addition to this model, $P$ has another stable model $\{s,q\}$ which
can be verified similarly by constructing the reduct and its least model.
\end{example}

The stable model semantics for programs with variables is obtained from
the semantics of ground programs by employing the notion of
\emph{Herbrand models}. The stable models of a  program with variables are
the stable models of the ground instantiation of the program where
variables are substituted by terms from the Herbrand universe of the
program (the ground terms built from constants and functions in the program).

\emph{Integrity constraints}, i.e., rules of the
form
\begin{equation}
\lparrow B_1,\ldots,B_m,\lpnot C_{1},\ldots,\lpnot C_n
\label{eq:icrule}
\end{equation}
are often useful for saying that a stable model containing $B_1,\ldots,B_m$
but none of $C_{1},\ldots, C_n$ is not acceptable.  These rules can be
encoded using ordinary rules\footnote{For example, by introducing two new
  atoms $f$ and $f'$ and a new rule $f' \lparrow \lpnot f', f$ and finally replacing every rule of the form (\ref{eq:icrule}) with one
having $f$ as its head.}.

\begin{example}
Consider program $P$ in Example~\ref{ex:smodel:intro} extended by
two integrity constraints
\begin{eqnarray*}
&& \leftarrow \lpnot p, s\\
&& \leftarrow r, \lpnot q, s
\end{eqnarray*}
This program has only one stable model $\{r,p\}$ as the other stable
model of $P$, $\{s,q\}$, does not satisfy the first integrity constraint
above. 
\end{example}

Integrity constraints are a powerful and simple technique for pruning
unwanted stable models as they cannot 
introduce new stable models but only can eliminate them. 
This means that for a program $P$ and  a set of integrity constraints $IC$,
if $S$ is a stable model of $P \cup IC$, then $S$ is a stable model of
$P$. 

\section{From Boolean Logic to Logic Programs} \label{sec:bool}

DES can be seen as a boolean function which takes as input a vector of bits
consisting of the plaintext and key and outputting a vector of bits (the
ciphertext). DES is specified using standard boolean operators (negation,
disjunction, conjunction, XOR) as well as boolean functions given as truth
tables.

In this section we discuss how to encode such boolean expressions using logic
programs. Here the goal is to achieve a compact and potentially
computationally efficient coding. We aim to exploit the special property of
the stable model semantics that everything is false unless otherwise stated.
This means that it is enough to consider only the conditions under which an
expression is true and let the default negation to handle the other case when
the expression is false.

Given a boolean expression $\varphi$ we provide a logic program $P_\varphi$
such that satisfying truth assignments of $\varphi$ and stable models of
$P_\varphi$ coincide. This can be done by introducing a new atom $p_\psi$ for
each subexpression $\psi$ of $\varphi$ and, according the intuition mentioned
above, by only giving rules stating all conditions
on its subexpressions under which $\psi$ is true.

\begin{table}
\centering
\caption{Mapping boolean expression to rules \label{table:BtoRules}}
\vspace{5mm}
$
\begin{array}{|l|l|} \hline
\mbox{Subexpression} & \mbox{Rules} \\ \hline
l_1 \land\cdots\land l_n & 
\begin{array}[t]{l}
p \lparrow \atom{l_1},\ldots,\atom{l_n} 
\end{array}
\\ \hline
l_1 \lor \cdots\lor l_n &
\begin{array}[t]{l}
p \lparrow \atom{l_1} \\
\vdots  \\
p \lparrow \atom{l_n} 
\end{array}
\\ \hline
\neg l &
\begin{array}[t]{l}
p \lparrow \lpnot \atom{l} \\
\end{array}
\\ \hline
l_1 \xor l_2 &
\begin{array}[t]{l}
p \lparrow \atom{l_1},\lpnot \atom{l_2} \\
p \lparrow \lpnot \atom{l_1}, \atom{l_2} \\
\end{array}
\\ \hline
\end{array}
$
\end{table}

In Table~\ref{table:BtoRules} we give the corresponding rules for different
kinds of subexpressions. We use the convention that we denote by $p$ the
corresponding new atom of the subexpression in question and by $\atom{l}$ the
new atom introduced for any further subexpression $l$. 

As a further optimization, note that it is not necessary to introduce a new
atom in the program for negated subexpressions '$\neg l$' as they can be
represented as '$\lpnot p_l$' in the program, a positive literal can be
represented as such, and an expression '$\lpnot \lpnot a$' as '$a$'.

For the rest of the original propositional atoms, which are not introduced as
abbreviations in the original boolean expression, the assumption about the
default negation is false because they can have any of the two truth
values. Therefore we encode this by introducing a new atom $\hat{a}$ for each
atomic subexpression $a$ and including two rules
\begin{eqnarray}
&& a \lparrow \lpnot \hat{a} \label{eq:choice}\\
&& \hat{a} \lparrow \lpnot a \nonumber
\end{eqnarray}
stating that either $a$ is in the stable model or $\hat{a}$ is in the model
(when $a$ is not there).

Now the satisfying truth assignments of $\varphi$ and the stable models
$P_\varphi$ correspond in the following sense:
\begin{enumerate}
\item Each stable model $S$ of $P_\varphi$ induces a truth assignment $T$
  where an atom $a$ is true in $T$ iff $a \in S$ and for
  each subexpression $\psi$ of $\varphi$, $\psi$ is true in $T$ iff the
  corresponding new atom $p_\psi$ is in $S$.
  
\item Each truth assignment $T$ induces a stable model
  $S$ of $P_\varphi$ such that for each subexpression $\psi$ of $\varphi$,
  $\psi$ is true in $T$ iff the corresponding new atom $p_\psi$ is in $S$.
\end{enumerate}
In order to consider stable models corresponding to assignments where $\varphi$
is true, one adds to $P_\varphi$ a rule
\[
\lparrow \lpnot p_{\varphi}
\]

Further constraints on boolean (sub)expressions can be encoded
similarly. In order to ensure that a given (sub)expression $\psi$ is true (respectively
false), it is enough to include to $P_\varphi$ the rules
\[
\begin{array}{ll}
\lparrow \lpnot p_\psi & \mbox{forces } \psi \mbox{ to be true} \\
\lparrow        p_\psi & \mbox{forces } \psi \mbox{ to be false}
\end{array}
\]
where $p_\psi$ is the new atom corresponding to $\psi$.  Notice that our
translation can be seen as first breaking the boolean expression to a set of
equivalences where new atoms are defined for each expression and then mapping
these equivalence to rules.

\begin{example}
\label{ex:equivalence}
Consider an expression $\varphi$
\[
(a \lor \neg b) \land (\neg a \xor b)
\]
It can be seen as a set of equivalences
\[
\{p_1 \eqv p_2 \land p_3, p_2 \eqv a \lor \neg b, p_3 \eqv (\neg a \xor b) \}
.
\]
Now the program $P_\varphi$ is
\[
\begin{array}[t]{l}
 p_1 \lparrow p_2, p_3 \\
 p_2 \lparrow  a \\
 p_2 \lparrow  \lpnot b \\
 p_3 \lparrow  \lpnot a,\lpnot b \\
 p_3 \lparrow  a, b \\
\end{array}
\hspace{4em}
\begin{array}[t]{l}
a \lparrow \lpnot \hat{a} \\
\hat{a} \lparrow \lpnot a  \\
b \lparrow \lpnot \hat{b} \\
\hat{b} \lparrow \lpnot b  
\end{array}
\]
For instance, the stable model $\{a,\hat{b},p_2\}$ of $P_\varphi$
corresponds to the truth assignment where the atom $a$ is true but $b$
is false.  If we want to have only models where $\varphi$ true, it is enough to add to $P_\varphi$ the rule
\[
\lparrow \lpnot p_1 \;\; .
\]
When this is done, the resulting program has two stable models: 
$\{a, b, p_1, p_2, p_3\} $ and 
$\{\hat{a}, \hat{b}, p_1, p_2, p_3\} $.
\end{example}
A boolean function given as a truth table can be represented using rules by
considering a disjunctive normal form representation of the function. This
means that we give the conditions under which the function obtains the value
true and provide for each such case a corresponding rule.

\begin{example}
\label{ex:table}
The function $f$ given by the table on the left hand side can be encoded by
the rules on its right. 
\[
\begin{array}[c]{|lll|l|}
\hline
x_1 & x_2 & x_3 & f \\
\hline
0 & 0 & 0 & 1 \\
0 & 0 & 1 & 0 \\
0 & 1 & 0 & 1 \\
0 & 1 & 1 & 0 \\
1 & 0 & 0 & 0 \\
1 & 0 & 1 & 0 \\
1 & 1 & 0 & 1 \\
1 & 1 & 1 & 0 \\
\hline
\end{array}
\hspace{3em}
\begin{array}[c]{l}
 f \lparrow \lpnot x_1, \lpnot x_2, \lpnot x_3 \\
 f \lparrow \lpnot x_1,  x_2, \lpnot x_3 \\
 f \lparrow  x_1,  x_2, \lpnot x_3 \\
\end{array}
\]
\end{example}

\section{The US Data Encryption Standard} \label{des}

For a complete description of DES see~\cite{DES}, \cite[Chap.12]{Schn-94}, or
\cite{Stinson98}.
DES is a block-cipher and its input is a 64 bit block of
\emph{plaintext} and a 64 bit \emph{key}, where every eighth bit
is a parity check bit that is stripped off before the encryption. So,
the actual key-size of DES is 56 bits. This key is used for
generating the \emph{round-keys}, 48 bit permuted subkeys of the
key. The output is a 64 bit block of \emph{ciphertext}.

\begin{figure*}
\centering
\includegraphics[width=\textwidth]{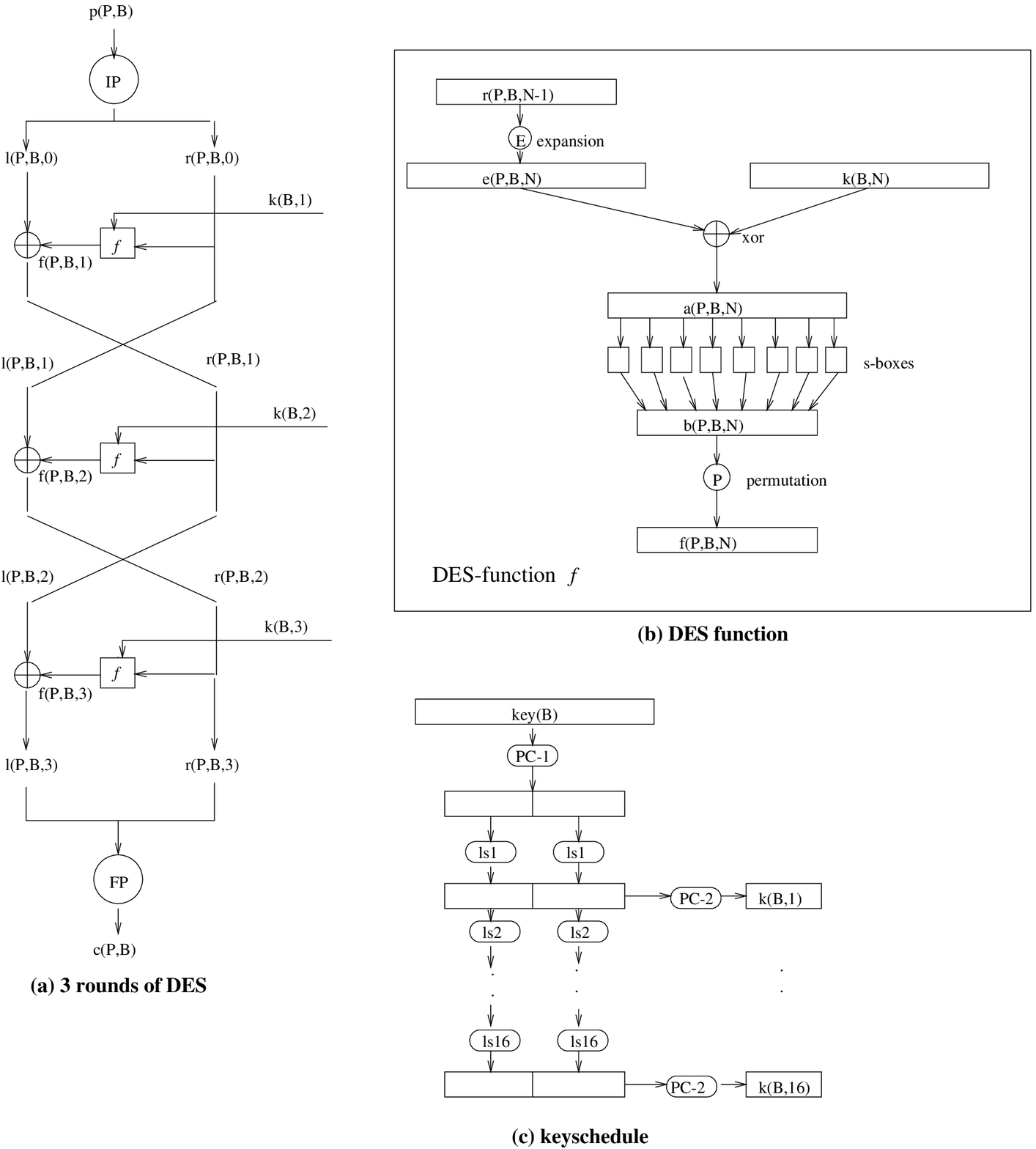}
\caption{The structure of DES} 
\label{DES:Structure}
\end{figure*}

The high level structure of DES is presented in Figure~\ref{DES:Structure}(a). Following Figure~\ref{DES:Structure}(a) top-down we see
that DES starts with  an initial permutation IP of the 64 bit block of
plaintext followed by a structure that is called a \emph{Feistel cipher} \cite{FNS-75}. 

The basic component of a Feistel cipher is called a \emph{round} and is
constituted by the following operations:
\begin{enumerate}
\item the input of 64 bits is divided into left and right parts;
\item the right half (32 bits), together with a round-key, is taken as input of a function $f$ (the round
function), which is described below;
\item the output of $f$ is XORed with the left
half and the result is a new right half;
\item the unaltered  old right half
becomes the new left half.
\end{enumerate}
These rounds can now be chained together and the complete DES contains
16 rounds (Figure~\ref{DES:Structure}(a) illustrates three
rounds). The strength of DES depends on the number of rounds: after 8
rounds a change in an input bit affects all output bits. In the end of
DES, the switching of left and right sides is omitted and the bits are
again permuted using the inversion of the initial permutation.

\paragraph{DES function $f$.} Inside the function $f$ (see Figure
\ref{DES:Structure}(b)) the 32 input bits are first expanded to 48 bits by
duplicating some of them. The expanded bit string is XORed with the round-key
given by the keyschedule described below. The resulting bits are input to 8
S-boxes, 6 bits for each box. The S-boxes are functions of six (binary)
variables. The output of every S-box consists of 4 bits. The resulting 32 bits
are finally permuted according to permutation P. The content of the boxes was
decided at the time DES was developed and they are the only non-linear part of
DES, hence the security of DES relies on them.

\paragraph{Keyschedule.} The keyschedule function takes as input the key and
provides as out a 48 bit round-key for each round of the Feistel cipher. 
The DES key is a vector of 64 bits, where every 8th bit is a
parity bit. First the parity bits are stripped off, then the keybits
are permuted according to the permutation PC-1. The result is divided
into two parts that are shifted to the left one or two positions
recursively, see Figure~\ref{DES:Structure}(c). After each shift the
bit string is again permuted (PC-2) in order to produce the
round-key.

\section{A Direct Encoding of DES} \label{direct}

We develop first a direct encoding of DES as a logic program.  It does not
contain any optimization and the idea is to keep the code simple and readable.
The code can be used for encryption or key search\footnote{We have been
  successful only for limited versions of DES where the number of rounds is
  less than 16 (the full version).} with several plaintext-ciphertext pairs
(the known plaintext attack).
 
The encoding is given as rules with variables. However, each variable has a
domain predicate in the body of the rule so that a set of ground
instances with exactly the same stable models is straightforward to
determine. 
The predicates contain variables $P$ indicating a
plaintext-ciphertext pair and $N$ for round. 
The corresponding domain predicates are $round(N)$ and $pair(P)$ 
which specify the relevant rounds and pairs, respectively. 
The total number of rounds is denoted by a constant $r$. For instance,
if we are considering a three round version of DES with two 
plaintext-ciphertext pairs, these domain predicates would be defined
using the facts:
\[
\begin{array}[t]{l}
round(0) \\
round(1) \\
\end{array}
\;\;\;\;
\begin{array}[t]{l}
round(2) \\
round(3) 
\end{array}
\;\;\;\;
\begin{array}[t]{l}
pair(1) \\
pair(2) 
\end{array}
\]

We describe first DES as used for encryption and then indicate changes
needed to be done, e.g., for key search. 
The plaintext is given as facts $p(P,B)$, where $B \in [1,64]$ gives the
number of the bit and 
$P$ indicates the pair in question. Note that only 
facts for true plaintext bits need to be written.
For instance, a set of facts 
\[
\{ p(1,1), p(1,2), \ldots, p(1,16)\}
\]
specifies that in the first plaintext-ciphertext pair, the plaintext bits
$1,\ldots,16$ are true and all other false. 

\subsection{Round Operations}

The rules which encode the round operations, i.e., the rules
that join the previous round to the next, are summarized in
Figure~\ref{fig:round}. They work as follows.

\def\spazio{}
\def\carat{}

\begin{figure*}
\centering
$
\begin{array}{ll}
\inlab{1} & 
permuted\_plaintext(P,B1) \lparrow ip(B,B1), p(P,B),\spazio {\carat pair}(P)\\
\inlab{2} & 
r(P,IB,0) \lparrow permuted\_plaintext(P,B1), renumber\_right(B1,IB),\spazio {\carat pair}(P), bit(B)\\
\inlab{3} & 
l(P,IB,0) \lparrow permuted\_plaintext(P,B1), renumber\_left(B1,IB),\spazio  {\carat pair}(P), bit(B)\\ 
\inlab{4} & 
l(P,B,N+1) \lparrow r(P,B,N), N+1<r,\spazio {\carat bit}(B), {\carat pair}(P), {\carat
  round}(N), {\carat round}(N+1)\\ 
\inlab{5} & 
r(P,B,N+1) \lparrow l(P,B,N), \lpnot f(P,B,N+1),  N+1<r,\spazio {\carat bit}(B), {\carat pair}(P), {\carat
  round}(N), {\carat round}(N+1)\\
\inlab{6} & 
r(P,B,N+1) \lparrow \lpnot l(P,B,N), f(P,B,N+1),  N+1<r,\spazio {\carat bit}(B), {\carat pair}(P), {\carat
  round}(N), {\carat round}(N+1)\\
\inlab{7} & 
r(P,B,r) \lparrow r(P,B,r-1),\spazio {\carat bit}(B), {\carat pair}(P)\\
\inlab{8} & 
l(P,B,r) \lparrow l(P,B,r-1), \lpnot f(P,B,r),\spazio {\carat bit}(B), {\carat pair}(P)\\
\inlab{9} & 
l(P,B,r) \lparrow \lpnot l(P,B,r-1), f(P,B,r), \spazio {\carat bit}(B), {\carat pair}(P)\\
\inlab{10} & 
unpermuted\_cipher(P,B1) \lparrow r(P,IB,r), renumber\_right(B1,IB), \spazio {\carat pair}(P)\\
\inlab{11} & 
unpermuted\_cipher(P,B1) \lparrow l(P,IB,r), renumber\_left(B1,IB),
\spazio {\carat pair}(P)\\ 
\inlab{12} & 
cipher(P,BC) \lparrow ip(BC,B1), unpermuted\_cipher(P,B1),\spazio {\carat pair}(P)\\
\end{array}
$
\caption{Round operations} \label{fig:round}

\end{figure*}

For the first round, the 64-bit block of plaintext is first permuted
according to the initial-permutation IP which is given as a set of facts
\[
\{ip(1,40), \ldots, ip(9,39) \}
\]
Using these facts the rule for $permuted\_plaintext$ \inlab{1} is easy to
express.

For each pair $P$, the bits are divided in two 32
bit parts and renumbered. The renumbering is used only to make the
description of the function $f$ easier to write and understand and 
it is done by dividing both halves (32 bits) 
into 8 groups with each 4 bits. The bits are numbered so that the
first digit represents the group and the second digit represents the
bit. For example, bit 32, is the second bit in the third group. This
renumbering is given as facts:
\[
\begin{array}{l}
renumber\_left(1,11)\\
renumber\_left(2,12)\\
renumber\_left(3,13)\\
renumber\_left(4,14)\\
renumber\_left(5,21)\\
\ldots \\
renumber\_left(32,84) \\
\end{array}
\;\;\;\;
\begin{array}{l}
renumber\_right(33,11)\\
renumber\_right(34,12)\\
renumber\_right(35,13)\\
renumber\_right(36,14)\\
renumber\_right(37,21)\\
\ldots \\
renumber\_right(64,84) \\
\end{array}
\]
These facts are used in the rules \refinlab{2} and \refinlab{3}
specifying the
right and left parts where the predicate $bit(B)$ is defined
using a set of facts giving the possible renumbered bits 
$11,12,13,14,21, \ldots, 84$. 

For each round $N+1$ and each plaintext-ciphertext pair $P$, the left
and right parts $l(P,B,N+1)$ and $r(P,B,N+1)$ can be defined in terms of
the previous parts and the result of the function $f$ as follows.
The right side is swapped to the left (\refinlab{4})
and the left side is XORed with the output of $f$ to
form the right side for the next round (\refinlab{5}--\refinlab{6}).

In the final round the switching of left and
right halves is omitted (\refinlab{7}--\refinlab{9}) but
the renumbering is undone and the final
permutation (\refinlab{10}--\refinlab{12}) is applied.

\subsection{Function $f$}\label{DES function}

For each round $N$ and for each pair $P$, the function $f$ takes as input the
32 bits of the right part of the previous round $r(P,B,N-1)$ and a 48 bit
round-key $k(B,N)$ and works as follows.  First every group of the right part
is expanded from 4 to 6 bits. For example, the rule
\begin{equation}
\begin{array}[b]{ll}
e(P,65,N) \lparrow  &r(P,64,N-1), round(N), \\
&round(N-1), pair(P)
\end{array}
\label{eq:expansion}
\end{equation}
means that the 4th bit in the 6th group becomes the 5th bit in the 6th
group.  The expanded bit string is XORed with the
key bits:
\[
\begin{array}{ll}
a(P,B,N)\lparrow& e(P,B,N), \lpnot k(B,N), \\
&round(N), N\not=0, pair(P), ebit(B)\\
a(P,B,N)\lparrow& \lpnot e(P,B,N), k(B,N), \\
&round(N), N\not=0, pair(P), ebit(B)
\end{array}
\]
where the predicate $ebit(B)$ is defined
using a set of facts giving the possible extended renumbered bits
$11,12,13,14,15,16,21, \ldots, 86$. 

The resulting groups of 6 bits are the input of their
respective S-boxes. The output of every S-box consists of 4 bits. If
we consider the output one bit at the time, the S-boxes can be seen as
truth tables. For example, if the input to the second S-box is 010101, it's
output is 0001. We can encode this behavior with the following rule:
\begin{displaymath}
\begin{array}{ll}
b(P,24,N) \lparrow& \lpnot a(P,21,N),a(P,22,N), \\
&\lpnot a(P,23,N),a(P,24,N),\\
&\lpnot a(P,25,N), a(P,26,N),\\
&round(N), N \not=0, pair(P).
\end{array}
\end{displaymath}
Once again, with the stable models semantics only rules that
imply true output bits are needed (see, Example~\ref{ex:table}). In this case, 
the output bits 1-3 are zeros, therefore no rules are needed for them.  
In the end of the DES function, the vector
of bits is permuted according to the permutation P. The rules for permutation
are similar to the ones in expansion.
 
For each round $N$, the keyschedule is given as a set of rules using the
key bit facts $key(K)$.  For example, the rule
\[k(11,1) \lparrow key(10), round(1) \]
specifies that in the first round the (renumbered) bit 11 of the round-key is
determined by the key bit 10. 
The stages presented in Figure \ref{DES:Structure}(c) and the renumbering is 
calculated beforehand in order to avoid some modulo arithmetic. This 
can be done because the keyschedule is independent of the plaintext to
be used. 

\subsection{Encryption and key search}

The encoding can be easily modified to solve many kinds of computational
problems related to DES by changing the way the plaintext, ciphertext
and the key are encoded. 

\begin{description}
\item[Encryption:] It is sufficient to give the true bits $B$ of the
plaintext as facts $p(P,B)$ for each pair $P$ and the true bits of the
key as facts $key(K)$. Now for each pair $P$, the true bits of the
encrypted ciphertext can be recovered as ground facts $cipher(P,B)$ in the unique
stable model of the encoding with the plaintext and key facts. 

\item[Decryption:] The true bits of the key are specified as facts $key(K)$, the
  ciphertext is given in the form 
\[
\begin{array}{ll}
\lparrow cipher(P,B) & \mbox{for 0-bits} \\
\lparrow \lpnot cipher(P,B) & \mbox{for 1-bits}
\end{array}
\]
  and the plaintext  by the
  rules of the form (\ref{eq:choice}) saying that one can choose the truth
  values of the ground atoms $p(P,B)$. Then the decrypted plaintext is given
  by the stable model of the encoding: for each true bit of the
  plaintext a ground fact $p(P,B)$ is in the model.

  Actually, DES is symmetric. This means that decryption is usually
  done the same way as encryption, using the key schedule in reverse
  order and the ciphertext in place of the plaintext.

  
\item[Known plaintext attack:] For this attack we assume that a certain
  number of pairs of plaintexts and the corresponding ciphertexts are available
  and that we want to recover the key. For each pair $P$, the true bits
  $B$ of the plaintext are given as facts $p(P,B)$, the ciphertext is
  given in the form
\[
\begin{array}{ll}
\lparrow cipher(P,B) & \mbox{for 0-bits} \\
\lparrow \lpnot cipher(P,B) & \mbox{for 1-bits}
\end{array}
\]
and the key is given by rules of the form (\ref{eq:choice})
\begin{eqnarray*}
&& key(k) \lparrow \lpnot \widehat{key}(k) \\
&& \widehat{key}(k) \lparrow \lpnot key(k) \nonumber
\end{eqnarray*}
specifying that the truth values of the ground atoms $key(k)$
corresponding to the key bits can be chosen. Then the stable models of
the resulting encoding correspond to the possible keys yielding the
ciphertext from the plaintext for each pair $P$. A key is given as
ground facts $key(K)$ in the corresponding stable model for all true key
bits.
\end{description}

\section{An Optimized Encoding of DES} \label{sec:optimized}

Massacci and Marraro~\cite{mass-marr-00-JAR} have devised an optimized
encoding of DES to SAT which is particularly effective when the plaintext and
the ciphertext are used in a known plaintext attack.  We show how to modify
this to work with logic programs.  We sketch here just the main ideas to make
the paper self-contained and refer to \cite{mass-marr-00-JAR} for further
details on the encoding.

The basic idea of the direct encoding is to represent each step of DES as 
a logic program, the more straightforward, the better. 
For the optimized encoding we start from a different direction and represent
DES as a logical circuit in which each operation is represented as a boolean
formula.

Then, for the operations that are repeated at each round (such as the round
function $f$) we
apply off-line some advanced CAD minimization techniques to squeeze their size
as much as possible. In particular in \cite{mass-marr-00-JAR} the CAD program
{\sf Espresso} \cite{Rude-Sang-87} has been used for minimizing the
representation of S-Boxes as Programmable Logic Arrays (PLAs). The PLA
representation is just a representation of boolean functions with disjunctions
of conjunctions.

This yields a notable squeeze in the size of the boolean formulae
representing the corresponding operations of the S-Boxes but is not
enough. The second important twist is that whenever possible, the program
``executes'' directly the DES operations on the propositional variables
representing the input bits. For instance, a permutation is not encoded into a
boolean formula, rather the program executes the permutation of the input bits
and provides as output the permuted propositional variables. 

The simplifying effect of this operation can be also explained as a form of
partial evaluation in the direct encoding of DES. Consider, for instance,
the logic program rule (\ref{eq:expansion}). The net
effect of the ``execution'' step is that $e(P,65,N+1)$ is replaced everywhere
by $r(P,64,N)$.

At the end of this process the encoder program {\sf def2fml} used in
 \cite{mass-marr-00-JAR} could output a minimized logic program corresponding
 to DES w.r.t.\ the direct encoding that we have described in the previous
 section using the rules we have given in the section on coding boolean formulae.

We can do more when the plaintext and the ciphertext are known, i.e.\ when we
want to perform a known plaintext attack. In particular, with a  boolean
representation we can perform a notable amount of linear reasoning (reasoning
using formulae with exclusive or). In \cite{mass-99-IJCAI} it is noted that
the presence of exclusive or is what makes the problem hard for
state-of-the-art SAT checkers and therefore its minimization is essential.

So, for the encoding we acquire the boolean values corresponding to plaintext
and ciphertext and preprocess the formula by applying exhaustively a set of
simplification rules aimed at eliminating redundancies:
\begin{enumerate}
\item Variables defined by atomic equivalences\footnote{We define an atomic
    equivalence as a formula of the form $V\eqv F$ where $V$ is a variable and
    $F$ is either another variable or a truth value.} are replaced by the
  corresponding values to reduce the number of variables in other formulae,
  and to introduce the truth values.
\item The propositional simplification rules listed in
  Figure~\ref{fig:valprop} are applied.
\end{enumerate}
The second step (propositional simplification) may introduce additional atomic
equivalences and therefore the overall simplification phase is repeated until
saturation is reached.

        \begin{figure}[t]
         \begin{center}
\begin{tabular}[t]{l}
      \begin{tabular}[t]{|l|l|}
        \hline
        Formula                       &  Simplification\\
        \hline
        $A \eqv X \wedge X$              &  $A\eqv X$\\
        $A \eqv X \wedge 0$              &  $A\eqv 0$\\
        $A \eqv X \wedge 1$              &  $A\eqv X$\\
        $A \eqv  X \wedge \overline{X}$   &  $A\eqv 0$\\
        $A \eqv  X \vee X$                &  $A\eqv X$\\
        $A \eqv  X \vee 0$                &  $A\eqv X$\\
        $A \eqv  X \vee 1$                &  $A\eqv 1$\\
        $A \eqv  X \vee \overline{X}$     &  $A\eqv 1$\\
        $A \eqv  X \oplus X$              &  $A\eqv 0$\\
        $A \eqv  X \oplus \overline{X}$   &  $A\eqv 1$\\
        $A \eqv  X \oplus 0$              &  $A\eqv X$\\
        $A \eqv  X \oplus 1$              &  $A \eqv
        \overline{X}$\\
\hline
      \end{tabular}
\\
\begin{tabular}[t]{|l|l|}
\hline
    Formula & Generated Equivalence\\
\hline
        $1 \eqv  A \wedge B$              &  $A\eqv 1$; $B
\eqv 1$\\
        $0 \eqv  A \vee B$                &  $A\eqv 0$; $B
\eqv 0$\\
        $0 \eqv  A \oplus B$              &  $A\eqv B$\\
        $1 \eqv  A \oplus B$              &  $A\eqv \overline{B}$\\
        $A \eqv  A \oplus B$              &  $B\eqv 0$\\
        $\overline{A} \eqv  A \oplus B$   &  $B\eqv 1$\\
        \hline
      \end{tabular}
        \end{tabular}
      \end{center}
      \caption{Simplification rules}         \label{fig:valprop}
    \end{figure}
 
Notice that such preprocessing, and in particular the operations involving
exclusive or, cannot be performed with a logic program representation (at
least with current technology).
   
The resulting formula is then translated into a logic program using a further
optimized translation w.r.t.\ that presented in the section on boolean encoding.
We can exploit the knowledge that the final formula we got has the form shown
in Figure~\ref{fig:encod:preprocess} (adapted from \cite{mass-marr-00-JAR}) and
translate it as shown in Figure~\ref{fig:encod:translate}. The variables $P$
and $N$ stand for the number of pair and rounds, according the format of the
direct encoding. The letter $b$ corresponds to a suitable ground value of the
bit number represented by the variable $B$ used in the direct encoding. 
Notice that the final output is a ground logic program so that $N$ and $P$ are
appropriately instantiated by the optimizing encoder.

\begin{figure}
  \begin{center}
\begin{displaymath}
\begin{array}{@{}l@{}}
r(P,b,3) \eqv \pm s(P,b'',1)\oplus s(P,b'',3)  \\
r(P,b,4) \eqv \pm s(P,b',2)\oplus s(P,b'',4)  \\
r(P,b,N) \eqv r(P,b',{N\!-\!2})\oplus s(P,b'',N), 5\leq\! N\!\leq r-4  \\
s(P,b,N)\eqv \bigvee_{b'} m(P,b',N), N=1\ldots r  \\
\pm s(P,b,r-1) \eqv r(P,b',r-5)\oplus s(P,b'',r-3)  \\
\pm s(P,b,r) \eqv r(P,b',r-4)\oplus s(P,b'',r-2)  \\
m(P,b,1) \eqv \bigwedge_{b'} \pm k(b',1)  \\
m(P,b,N) \eqv \bigwedge_{b'} x(P,b',N), 2\leq N\leq r-1  \\
m(P,b,r) \eqv \bigwedge_{b'} \pm k(b',r)  \\
x(P,b,2) \eqv \pm s(P,b',1)\oplus k(b'',2)  \\
x(P,b,3) \eqv \pm s(P,b',2)\oplus k(b'',3)  \\
x(P,b,N) \eqv r(P,b',i-1)\oplus k(b'',N), 4\leq N\leq r-3  \\
x(P,b,r-2) \eqv \pm s(P,b',r-1)\oplus k(b'',r-1)  \\
x(P,b,r-1) \eqv \pm s(b',r)\oplus k(b'',r) 
\end{array}
\end{displaymath}
\caption{Simplified DES formulae for $r$ rounds with known plaintext and ciphertext} \label{fig:encod:preprocess}
\end{center}
\end{figure}

\begin{figure*}
\centering
    $
\begin{array}[t]{@{}l@{}}
r(P,b,3)  \lparrow  \pm s(P,b',1), \lpnot s(P,b'',3)  \\
r(P,b,3)  \lparrow  \lpnot \pm s(P,b',1), s(P,b'',3)  \\
r(P,b,4)  \lparrow  \pm s(P,b',2), \lpnot s(P,b'',4)  \\
r(P,b,4)  \lparrow  \pm \lpnot s(P,b',2), s(P,b'',4)  \\
r(P,b,N)  \lparrow  r(P,b',N-2), \lpnot s(P,b'',N), 5\leq N\leq r-4  \\
r(P,b,N)  \lparrow  \lpnot r(P,b',N-2), s(P,b'',N), 5\leq N \leq r-4  \\
s(P,b,N) \lparrow   m(P,b',N), 1\leq N\leq r \mbox{ and } 1\leq b'\leq n_N   \\
s(P,b,r-1) \lparrow  \pm r(P,b',r-5), \lpnot s(P,b'',r-3) \\
s(P,b,r-1) \lparrow  \lpnot \pm r(P,b',r-5), s(P,b'',r-3)  \\
s(P,b,r) \lparrow  \pm r(P,b',r-4), \lpnot s(P,b'',r-2)  \\
s(P,b,r) \lparrow  \lpnot \pm r(P,b'',r-4), s(P,b'',r-2)  \\
m(P,b,1) \lparrow  \pm k(b',1), \ldots \pm k(b'',1)  \\
m(P,b,N) \lparrow  x(P,b',N)_1, \ldots, x(P,b'',N)_{n_N}, 2\leq N \leq r-1  \\
m(P,b,r) \lparrow  \pm k(b',r), \ldots \pm k(b'',r)  \\
\end{array}
\begin{array}[t]{@{}l@{}}
x(P,b,2)  \lparrow  \pm s(P,b',1), \lpnot k(b'',2) \\
x(P,b,2)  \lparrow  \pm \lpnot s(P,b',1), k(b'',2) \\
x(P,b,3)  \lparrow  \pm s(P,b',2), \lpnot k(b'',3) \\
x(P,b,3)  \lparrow  \pm \lpnot s(P,b',2), k(b'',3) \\
x(P,b,N)  \lparrow  r(P,b',N-1), \lpnot k(b'',N), 4\leq N\leq r-3  \\
x(P,b,N)  \lparrow  \lpnot r(P,b',N-1), k(b'',N), 4\leq N \leq r-3  \\
x(P,b,r-2)  \lparrow  \pm s(P,b',r-1), \lpnot k(b'',r-1)  \\
x(P,b,r-2)  \lparrow  \pm \lpnot s(P,b',r-1), k(b'',r-1)  \\
x(P,b,r-1)  \lparrow  \pm s(P,b',r), \lpnot k(b'',r) \\
x(P,b,r-1)  \lparrow  \pm \lpnot s(P,b',r), k(b'',r)
\end{array}
$
\caption{Optimized logic program for DES with $r$ rounds} \label{fig:encod:translate}
\end{figure*}

Notice that the translation of the formula is done piecewise: each equivalence
is translated in a suitable number of rules: we use one rule for conjunctions,
two rules for XORs, and many rules of disjunctions (as many as there are
disjuncts).
The trick is that we only encode one direction of the the equivalence
exploiting the property of logic programs that ``everything is false by
default''. In this way we have only to specify when a boolean formula may be
true.

However, this is still not sufficient because the translation as sketched is
not faithful: we might have more than one ``definition'' of the same atom,
i.e. one or more formulae of the form $A\eqv \phi_1$ for the same atom $A$.

If we left it that way, there would not be a one-one correspondence between
stable models and propositional truth assignments. We would have more models
than due. So we need a further twist to cope
with atoms that are defined (are on the left of the equivalence sign in
Figure~\ref{fig:encod:preprocess}) two or more times. Suppose that we have a
set of formulae of the form:
\[
a \eqv \varphi_1, \ldots, a \eqv \varphi_n .
\]
and that \translate{a \eqv  \varphi_i} denotes the fragment of the logic
program translating the boolean formula $a \eqv  \varphi_i$ according the
rules we have used in Table~\ref{table:BtoRules} and Figure~\ref{fig:encod:translate}.

We translate this set of formulae as follows:
\[
\begin{array}[t]{l@{\hspace{4ex}}l}
\mbox{Boolean formula} & \mbox{Logic program} \\[1ex]
a \eqv  \varphi_1 &  \translate{a \eqv  \varphi_1} \\[1ex]
a \eqv  \varphi_2 & a \lparrow  a_2 \\
               & \lparrow a,\lpnot a_2\\
               & \translate{a_2 \eqv  \varphi_2} \\
\vdots         & \vdots \\
a \eqv  \varphi_n & a \lparrow  a_n \\
               & \lparrow a,\lpnot a_n\\
               & \translate{a_n \eqv  \varphi_n}
\end{array}
\]
One may check that this is a faithful translation of the corresponding boolean
formulae. The intuitive explanation is simply that the boolean set of formulae,
read conjunctively, just says that all $\varphi_i$ must have the
same value and this value must also be assigned to $a$. The first rule chooses
a value, say $\varphi_1$ and assign it to $a$ as in the standard encoding.
The rest of the construction assigns the value of $\varphi_i$ to a new atom
$a_i$ and then specifies that $a$ is true when $a_i$ is true and that $a_i$
cannot be false when $a$ is true.

Then we add the rules (\ref{eq:choice}) saying that one can choose  the truth
values of the atoms corresponding to key bits, as we do for the direct encoding, and we
are done.

\section{Experiments}

We study the computational properties of the two logic program encodings
of DES by using them for key search in a known plaintext attack for a
limited form of DES running a given number of rounds.  For each number
of rounds and pairs of plaintext-ciphertext blocks we perform
50 key searches using different randomly generated plaintexts and report
the mean of the running time and of the size of the search tree.  The tests
were run under Linux 2.2.12 on 450 MHz Pentium III computers. The
encodings and test cases are available at
\texttt{http://www.tcs.hut.fi/Software/\linebreak[3]smodels/\linebreak[3]tests/\linebreak[3]des.html}.

Table~\ref{table:smodels} reports the data on \SMODELS's performance.
The running times do not include preprocessing. For the direct encoding
(Dir.)\ preprocessing consists of parsing and grounding of the rules
which is done by the standard \SMODELS\ parser \lparse. This takes only
few seconds even for the largest examples.  For the optimized encoding
(Opt.)\ preprocessing is more involved as explained in the previous
section. It includes off-line minimization of boolean functions used in
DES, partial evaluating the DES description, simplifying it using the
known plaintexts-ciphertext pairs, transforming the resulting boolean
formula to a set of ground logic program rules as well as parsing the
rules into the internal format of \smodels. 
Hence,  in both cases preprocessing produces 
a ground program parsed into the internal format of \smodels. 
Table~\ref{table:smodels} gives the average running time and search
space size for \smodels\ (version 2.25 with \texttt{-backjump} option)
to find a stable model (a key) for such a ground program. 
%
Entries marked with '\putchk' are cases where the set of 50 key searches
could not be completed because the running time for each key search
extended several CPU hours. 

Both encodings have a reasonable performance (although it should be
noted that special purpose methods and hardware are able to perform
known plaintext attacks successfully even to the full DES).  The direct
encoding does not seem to be able to propagate the information from the
known plaintext-ciphertext pairs as efficiently as the preprocessing
techniques in the optimized encoding.  The search heuristics of
\smodels\ yields a rather stable performance on these DES examples
except for the optimized encoding  with three rounds and two blocks where
there are three orders of magnitude differences in the minimal and
maximal observed running times and search space sizes.

We compare the performance of \SMODELS\ to that of a SAT-checker which has
been customized and tuned for the optimized SAT-encoding of DES described in
\cite{mass-99-IJCAI,mass-marr-00-JAR}. This SAT-checker, based on \RELSAT\ by Bayardo and
Schrag \shortcite{Baya-Schr-97}, clearly outperforms state-of-the-art SAT-checkers
on DES encodings~\cite{mass-marr-00-JAR}.

Table~\ref{table:relsat} reports the data on \RELSAT.  The data does not
include preprocessing which in this case is similar to that of the
optimized logic program encoding with the addition that it includes also
the transformation of the optimized DES description (a boolean formula)
to a compact conjunctive normal form (CNF) representation.
Table~\ref{table:relsat} presents the average running time and search
space size for \RELSAT\ to find a propositional model (a key) for this
CNF formula.

From this preliminary analysis one can say that the usage of stable models as
computational paradigm to be used in practice does not score at all badly for
such an industrial application. 

\begin{table}
\caption{\SMODELS\ on DES} \label{table:smodels}

\centering

\begin{tabular}{|@{\hspace{1pt}}r|r||r|r|r|r|}
\hline
\multicolumn{6}{|c|}{\SMODELS\ } \\
\hline
Rounds & Blocks & \multicolumn{2}{c|}{Time (s)} &
\multicolumn{2}{c|}{Branches} \\ 
       &        & (Dir.) & (Opt.)   & (Dir.) & (Opt.) \\
\hline
1 & 1 & 0.3& 0.07 & 155 & 28 \\
1 & 2 & 1.6 & 0.06 & 372 & 18 \\
1 & 4 & 2.2 & 0.1 & 179& 16 \\
1 & 8 & 5.8 & 0.2 & 200 & 16 \\
\hline
2 & 1 & 1.2 & 0.1 & 151& 9 \\
2 & 2 & 1.7  & 0.1 & 98 & 8\\
2 & 4 & 2.0 & 0.2 & 51 & 8\\
2 & 8 & 3.6  & 0.4  & 39 & 8 \\
\hline
3 & 1 & \putchk &  230       & \putchk &     699   \\
3 & 2 & 640 &  8900 & 20672 & 6000 \\
3 & 4 & 1400 & 190 & 14709 & 29 \\
3 & 8 & 3500 & 48 & 18612 & 8\\
\hline
\end{tabular}
\end{table}

\begin{table}
\caption{\RELSAT\ on DES} \label{table:relsat}
\centering
\begin{tabular}{|r|r||r|r|}
\hline
\multicolumn{4}{|c|}{\RELSAT\ with learning factor 5} \\
\hline
Rounds & Blocks & Time (s) & Branches \\
\hline
1 & 1 & 0.02 & 32\\
1 & 2 & 0.1 & 100\\
1 & 4 & 0.2 & 107 \\
1 & 8 & 0.4 & 87\\
\hline
2 & 1 & 0.2 & 283\\
2 & 2 & 0.2 & 106\\
2 & 4 & 0.3 & 70\\
2 & 8 & 0.7 & 56\\
\hline
3 & 1 & \putchk & \putchk \\
3 & 2 & 920 & 141291\\
3 & 4 & 110 & 14419 \\
3 & 8 & 100 & 5483 \\
\hline
\end{tabular}
\end{table}

\section{Conclusions}

We believe that DES provides an interesting benchmark problem for
nonmonotonic reasoning systems because
(i)~it supplies practically inexhaustible number of industrial
relevant test cases,
(ii)~the encoding of DES using normal logic programs with
the stable model semantics is easy to understand, and
(iii)~test cases are obtained for many nonmonotonic formalisms which
contain this subclass of logic programs as a special case. 
We have developed a direct encoding and an
optimized one extending the work of Massacci and Marraro. We have also tested the
computational performance of the encodings using the \SMODELS\ system. 

As DES is basically a boolean function, its
encoding does not require any particular nonmonotonic
constructs. In our encoding we have used default negation in a
straightforward way (everything is false unless otherwise stated), to obtain a
much leaner encoding than those obtained by encoding DES as a SAT formula
(where both ways of the equivalence are needed).
The resulting encodings are acyclic sets of 
rules which are compact but fairly simple to write and understand. It
seems that they are more easier to understand than corresponding
encodings of DES using CNF clauses which is the
typical input format for current state-of-the-art SAT-checkers.  
Given that DES key search is a natural boolean satisfiability problem,
it is somewhat surprising that our encodings are competitive when compared
to state-of-the-art SAT-checkers and even to a tuned and customized
SAT-checker working on an optimized SAT-encoding of DES.  
We think that the success can be accounted for by the compactness of the logic
program encoding and the search methods and pruning techniques employed in the
\SMODELS\ system.  

In order to obtain a deeper understanding of the relative strengths of
SAT-checkers and stable model implementations, an interesting comparison would
be to map the stable model finding problem of DES key search directly to a
satisfiability problem and use a state-of-the-art SAT-checker to solve the
resulting problem. As our encodings are acyclic programs, the reduction
could be done using, e.g., a completion approach~\cite{Fages94}.



\end{document}